\documentclass[conference, a4paper]{template/IEEEtran-modified}
\pdfoutput=1

\IEEEoverridecommandlockouts

\usepackage{cite}       

\ifx\pdfoutput\undefined
	\usepackage{graphicx}
\else
	\usepackage[pdftex]{graphicx}
\fi

\graphicspath{{figures/}}

\usepackage{url}        

\usepackage{amsmath}    
\usepackage[dvipsnames]{xcolor}
\usepackage{tikz}
\usepackage{booktabs}
\usepackage{amssymb}
\usepackage{caption}
\usepackage{subcaption}
\usepackage[T1]{fontenc}

\DeclareMathOperator*{\argmin}{arg\,min}
\DeclareMathOperator*{\softmax}{softmax}

\usepackage{listings}

\definecolor{stringgreen}{RGB}{50,220,15}

\newcommand{\specialcell}[2][c]{%
  \begin{tabular}[#1]{@{}l@{}}#2\end{tabular}}

\begin{document}


\title{A Tour of TensorFlow}


\author{
\authorblockN{Peter Goldsborough}
\authorblockA{Fakult\"{a}t f\"{u}r Informatik\\Technische Universit\"{a}t M\"{u}nchen\\
Email: peter.goldsborough@in.tum.de}
}


\specialpapernotice{Proseminar Data Mining}


\maketitle




\begin{abstract}

  Deep learning is a branch of artificial intelligence employing deep neural
  network architectures that has significantly advanced the state-of-the-art in
  computer vision, speech recognition, natural language processing and other
  domains. In November 2015, Google released \emph{TensorFlow}, an open source
  deep learning software library for defining, training and deploying machine
  learning models. In this paper, we review TensorFlow and put it in context of
  modern deep learning concepts and software. We discuss its basic computational
  paradigms and distributed execution model, its programming interface as well
  as accompanying visualization toolkits. We then compare TensorFlow to
  alternative libraries such as Theano, Torch or Caffe on a qualitative as well
  as quantitative basis and finally comment on observed use-cases of TensorFlow
  in academia and industry.\vspace{0.5cm}

\end{abstract}



\begin{keywords}
  Artificial Intelligence, Machine Learning, Neural Networks, Distributed
  Computing, Open source software, Software packages
\end{keywords}


\section{Introduction}

Modern artificial intelligence systems and machine learning algorithms have
revolutionized approaches to scientific and technological challenges in a
variety of fields. We can observe remarkable improvements in the quality of
state-of-the-art computer vision, natural language processing, speech
recognition and other techniques. Moreover, the benefits of recent breakthroughs
have trickled down to the individual, improving everyday life in numerous
ways. Personalized digital assistants, recommendations on e-commerce platforms,
financial fraud detection, customized web search results and social network
feeds as well as novel discoveries in genomics have all been improved, if not
enabled, by current machine learning methods.

A particular branch of machine learning, \emph{deep learning}, has proven
especially effective in recent years. Deep learning is a family of
representation learning algorithms employing complex neural network
architectures with a high number of hidden layers, each composed of simple but
non-linear transformations to the input data. Given enough such transformation
modules, very complex functions may be modeled to solve classification,
regression, transcription and numerous other learning tasks \cite{nature2015}.

It is noteworthy that the rise in popularity of deep learning can be traced back
to only the last few years, enabled primarily by the greater availability of
large data sets, containing more training examples; the efficient use of
graphical processing units (GPUs) and massively parallel commodity hardware to
train deep learning models on these equally massive data sets as well as the
discovery of new methods such as the \emph{rectified linear unit} (ReLU)
activation function or \emph{dropout} as a regularization technique\cite{relu,
  dropout, nature2015, rampasek}.

While deep learning algorithms and individual architectural components such as
representation transformations, activation functions or regularization methods
may initially be expressed in mathematical notation, they must eventually be
transcribed into a computer program for real world usage. For this purpose,
there exist a number of open source as well as commercial machine learning
software libraries and frameworks. Among these are Theano \cite{theano}, Torch
\cite{torch}, scikit-learn \cite{scikit} and many more, which we review in
further detail in Section II of this paper. In November 2015, this list was
extended by \emph{TensorFlow}, a novel machine learning software library
released by Google \cite{tensorflow}. As per the initial publication, TensorFlow
aims to be ``an interface for expressing machine learning algorithms'' in
``large-scale [\dots] on heterogeneous distributed systems'' \cite{tensorflow}.

The remainder of this paper aims to give a thorough review of TensorFlow and put
it in context of the current state of machine learning. In detail, the paper is
further structured as follows. Section \ref{sec:history} will provide a brief
overview and history of machine learning software libraries, listing but not
comparing projects similar to TensorFlow. Subsequently, Section \ref{sec:model}
discusses in depth the computational paradigms underlying TensorFlow. In Section
\ref{sec:code} we explain the current programming interface in the various
supported languages. To inspect and debug models expressed in TensorFlow, there
exist powerful visualization tools, which we examine in Section
\ref{sec:visual}. Section \ref{sec:comp} then gives a comparison of TensorFlow
and alternative deep learning libraries on a qualitative as well as quantitative
basis. Before concluding our review in Section \ref{sec:conclusion}, Section
\ref{sec:uses} studies current real world use cases of TensorFlow in literature
and industry.


\section{History of Machine Learning Libraries}\label{sec:history}

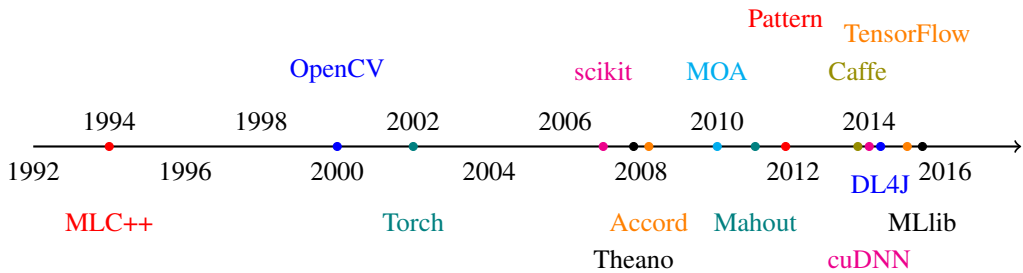
\begin{figure*}[t!]
\centering
  \begin{tikzpicture}
  \draw [thick, ->] (0, 0) -- (13, 0);

  \newcounter{y}
  \setcounter{y}{1992}
  \foreach \i in {0, ..., 12} {
    \draw (\i, {(Mod(\i, 2) - 0.5)/1.5}) node {\they};
    \stepcounter{y}
    \stepcounter{y}
  }

  \foreach \time/\lib/\a/\b/\c in {%
    1994/MLC++/-1/0/red,
    2000/OpenCV/+1/0/blue,
    2002/Torch/-1/0/teal,
    2007/scikit/+1/0/magenta,
    2008/Accord/-1/+0.1/orange,
    2008/Theano/-1.5/-0.1/black,
    2010/MOA/+1/0/cyan,
    2011/Mahout/-1/0/teal,
    2012/Pattern/+1.7/-0.1/red,
    2014/DL4J/-0.5/0.15/blue,
    2014/Caffe/+1/-0.15/olive,
    2014/cuDNN/-1.5/0/magenta,
    2015/TensorFlow/+1.5/0/orange,
    2015/MLlib/-1/0.2/black%
    } {
    \draw [\c] ({(\time - 1992)/2 + \b}, \a) node [align=center] {\lib};
    \filldraw [\c] ({(\time - 1992)/2 + \b}, 0) circle [radius=1.5pt];
  }

  \end{tikzpicture}
\caption{A timeline showing the release of machine-learning libraries discussed
  in section I in the last 25 years.}
\label{fig:timeline}
\end{figure*}

In this section, we aim to give a brief overview and key milestones in the
history of machine learning software libraries. We begin with a review of
libraries suitable for a wide range of machine learning and data analysis
purposes, reaching back more than 20 years. We then perform a more focused study
of recent programming frameworks suited especially to the task of deep
learning. Figure \ref{fig:timeline} visualizes this section in a timeline. We
wish to emphasize that this section does in no way compare TensorFlow, as we
have dedicated Section \ref{sec:comp} to this specific purpose.

\subsection{General Machine Learning}\label{sec:history-general}

In the following paragraphs we list and briefly review a small set of
\emph{general machine learning libraries} in chronological order. With
\emph{general}, we mean to describe any particular library whose common
use cases in the machine learning and data science community include \emph{but
  are not limited to} deep learning. As such, these libraries may be used for
statistical analysis, clustering, dimensionality reduction, structured
prediction, anomaly detection, shallow (as opposed to deep) neural networks and
other tasks.

We begin our review with a library published 21 years before TensorFlow:
\emph{MLC++} \cite{mlcpp}. MLC++ is a software library developed in the C++
programming language providing algorithms alongside a comparison framework for a
number of data mining, statistical analysis as well as pattern recognition
techniques. It was originally developed at Stanford University in 1994 and is
now owned and maintained by Silicon Graphics, Inc
(SGI\footnote{https://www.sgi.com/tech/mlc/}). To the best of our knowledge,
MLC++ is the oldest machine learning library still available today.

Following MLC++ in the chronological order,
\emph{OpenCV}\footnote{http://opencv.org} (\textbf{O}pen
\textbf{C}omputer \textbf{V}ision) was released in the year 2000 by Bradski et
al. \cite{opencv}. It is aimed primarily at solving learning tasks in the field
of computer vision and image recognition, including a collection of algorithms
for face recognition, object identification, 3D-model extraction and other
purposes. It is released under a BSD license and provides interfaces in multiple
programming languages such as C++, Python and MATLAB.

Another machine learning library we wish to mention is
\emph{scikit-learn}\footnote{http://scikit-learn.org/stable/} \cite{scikit}. The
scikit-learn project was originally developed by David Cournapeu as part of the
Google Summer of Code program\footnote{https://summerofcode.withgoogle.com} in
2008. It is an open source machine learning library written in Python, on top of
the NumPy, SciPy and matplotlib frameworks. It is useful for a large class of
both supervised and unsupervised learning problems.

The \emph{Accord.NET}\footnote{http://accord-framework.net/index.html} library
stands apart from the aforementioned examples in that it is written in the C\#
(``C Sharp'') programming language. Released in 2008, it is composed not only of
a variety of machine learning algorithms, but also signal processing modules for
speech and image recognition \cite{accord}.

\emph{Massive Online Analysis}\footnote{http://moa.cms.waikato.ac.nz} (MOA) is
an open source framework for online and offline analysis of massive, potentially
infinite, data \emph{streams}. MOA includes a variety of tools for
classification, regression, recommender systems and other disciplines. It is
written in the Java programming language and maintained by staff of the
University of Waikato, New Zealand. It was conceived in 2010 \cite{moa}.

The \emph{Mahout}\footnote{http://mahout.apache.org} project, part of Apache
Software Foundation\footnote{http://www.apache.org}, is a Java programming
environment for scalable machine learning applications, built on top of the
Apache Hadoop\footnote{http://hadoop.apache.org} platform. It allows for
analysis of large datasets distributed in the Hadoop Distributed File System
(HDFS) using the \emph{MapReduce} programming paradigm. Mahout provides
machine learning algorithms for classification, clustering and filtering.

\emph{Pattern}\footnote{http://www.clips.ua.ac.be/pages/pattern} is a Python
machine learning module we include in our list due to its rich set of
\emph{web mining} facilities. It comprises not only general machine learning
algorithms (e.g. clustering, classification or nearest neighbor search) and
natural language processing methods (e.g. n-gram search or sentiment analysis),
but also a web crawler that can, for example, fetch Tweets or Wikipedia entries,
facilitating quick data analysis on these sources. It was published by the
University of Antwerp in 2012 and is open source.

Lastly, \emph{Spark MLlib}\footnote{http://spark.apache.org/mllib} is an
open source machine learning and data analysis platform released in 2015 and
built on top of the Apache Spark\footnote{http://spark.apache.org/} project
\cite{spark}, a fast cluster computing system. Similar to Apache Mahout, it
supports processing of large scale \emph{distributed} datasets and training of
machine learning models across a cluster of commodity hardware. For this, it
includes classification, regression, clustering and other machine learning
algorithms \cite{mllib}.

\subsection{Deep Learning}\label{sec:history-dl}

While the software libraries mentioned in the previous section are useful for a
great variety of different machine learning and statistical analysis tasks, the
following paragraphs list software frameworks especially effective in training
deep learning models.

The first and oldest framework in our list suited to the development and
training of deep neural networks is \emph{Torch}\footnote{http://torch.ch},
released already in 2002 \cite{torch}. Torch consisted originally of a pure C++
implementation and interface. Today, its core is implemented in C/CUDA while it
exposes an interface in the Lua\footnote{https://www.lua.org} scripting
language. For this, Torch makes use of a LuaJIT (just-in-time) compiler to
connect Lua routines to the underlying C implementations. It includes, inter
alia, numerical optimization routines, neural network models as well as
general purpose n-dimensional array (tensor) objects.

\emph{Theano}\footnote{http://deeplearning.net/software/theano/}, released in
2008 \cite{theano}, is another noteworthy deep learning library. We note that
while Theano enjoys greatest popularity among the machine learning community, it
is, in essence, not a machine learning library at all. Rather, it is a
programming framework that allows users to declare mathematical expressions
\emph{symbolically}, as computational graphs. These are then optimized,
eventually compiled and finally executed on either CPU or GPU devices. As such,
\cite{theano} labels Theano a ``mathematical compiler''.

\emph{Caffe}\footnote{http://caffe.berkeleyvision.org} is an open source
deep learning library maintained by the Berkeley Vision and Learning Center
(BVLC). It was released in 2014 under a BSD-License \cite{caffe}. Caffe is
implemented in C++ and uses neural network layers as its basic computational
building blocks (as opposed to Theano and others, where the user must define
individual mathematical operations making up layers). A deep learning model,
consisting of many such layers, is stored in the Google Protocol Buffer
format. While models can be defined manually in this Protocol Buffer
``language'', there exist bindings to Python and MATLAB to generate them
programmatically. Caffe is especially well suited to the development and
training of \emph{convolutional neural networks} (CNNs or ConvNets), used
extensively in the domain of image recognition.

While the aforementioned machine learning frameworks allowed for the definition
of deep learning models in Python, MATLAB and Lua, the
\emph{Deeplearning4J}\footnote{http://deeplearning4j.org} (DL4J) library enables
also the Java programmer to create deep neural networks. DL4J includes
functionality to create Restricted Boltzmann machines, convolutional and
recurrent neural networks, deep belief networks and other types of deep learning
models. Moreover, DL4J enables horizontal scalability using distributed
computing platforms such as Apache Hadoop or Spark. It was released in 2014 by
Adam Gibson under an Apache 2.0 open source license.

Lastly, we add the NVIDIA Deep Learning
SDK\footnote{https://developer.nvidia.com/deep-learning-software} to to this
list. Its main goal is to maximize the performance of deep learning algorithms
on (NVIDIA) GPUs. The SDK consists of three core modules. The first,
\emph{cuDNN}, provides high performance GPU implementations for deep learning
algorithms such as convolutions, activation functions and tensor
transformations. The second is a linear algebra library, \emph{cuBLAS}, enabling
GPU-accelerated mathematical operations on n-dimensional arrays. Lastly,
\emph{cuSPARSE} includes a set of routines for \emph{sparse} matrices tuned for
high efficiency on GPUs. While it is possible to program in these libraries
directly, there exist also bindings to other deep learning libraries, such as
Torch\footnote{https://github.com/soumith/cudnn.torch}.


\section{The TensorFlow Programming Model}\label{sec:model}

In this section we provide an in-depth discussion of the abstract computational
principles underlying the TensorFlow software library. We begin with a thorough
examination of the basic structural and architectural decisions made by the
TensorFlow development team and explain how machine learning algorithms may be
expressed in its dataflow graph language. Subsequently, we study TensorFlow's
execution model and provide insight into the way TensorFlow graphs are assigned
to available hardware units in a local as well as distributed environment. Then,
we investigate the various optimizations incorporated into TensorFlow, targeted
at improving both software and hardware efficiency. Lastly, we list extensions
to the basic programming model that aid the user in both computational as well
as logistical aspects of training a machine learning model with TensorFlow.

\subsection{Computational Graph Architecture}\label{sec:model-graphs}

\begin{figure}
  \begin{tikzpicture}
    \draw [very thick] (1, 2) circle [radius=0.5cm] node {$z$};

    \draw [very thick] (0, 0) circle [radius=0.5cm] node {$x$};
    \draw [very thick] (2, 0) circle [radius=0.5cm] node {$y$};

    \draw [very thick, -stealth] (0, 0.5) -- ++(60:1.29);
    \draw [very thick, -stealth] (2, 0.5) -- ++(120:1.29);

    \draw (1, 1.2) node {$+$};

    \draw (1, -1) node {$z = x + y$};
  \end{tikzpicture}
  \hspace{0.2cm}
  \begin{tikzpicture}
    \draw [very thick] (1, 2) circle [radius=0.5cm] node {$\mathbf{x^\top w}$};

    \draw [very thick] (0, 0) circle [radius=0.5cm] node {$\mathbf{x}$};
    \draw [very thick] (2, 0) circle [radius=0.5cm] node {$\mathbf{w}$};

    \draw [very thick, -stealth] (0, 0.5) -- ++(60:1.29);
    \draw [very thick, -stealth] (2, 0.5) -- ++(120:1.29);

    \draw (1, 1.2) node {\texttt{dot}};

    \draw [very thick] (3.5, 0) circle [radius=0.5cm] node {$b$};

    \draw [very thick] (3.5, 2) circle [radius=0.5cm] node {$z$};

    \draw (3, 1.42) node {$+$};

    \draw [very thick, -stealth] (3.5, 0.52) -- (3.5, 1.48);
    \draw [very thick, -stealth] (1.52, 2) -- (2.98, 2);

    \draw [very thick] (2.2, 3.5) circle [radius=0.5] node {$\hat{y}$};

    \draw (2.2, 2.8) node {$\sigma$};

    \draw [very thick, -stealth] (3.5, 2.52) -- ++(140:1.1cm);

    \draw (2, -1) node {$\hat{y} = \sigma(\mathbf{x}^\top \mathbf{w} + b)$};
  \end{tikzpicture}
  \caption{Examples of computational graphs. The left graph displays a very
    simple computation, consisting of just an addition of the two input
    variables $x$ and $y$. In this case, $z$ is the result of the operation $+$,
    as the annotation suggests. The right graph gives a more complex example of
    computing a logistic regression variable $\hat{y}$ in for some example
    vector $\mathbf{x}$, weight vector $\mathbf{w}$ as well as a scalar bias
    $b$. As shown in the graph, $\hat{y}$ is the result of the \emph{sigmoid} or
    \emph{logistic} function $\sigma$.}
  \label{fig:graphs}
\end{figure}
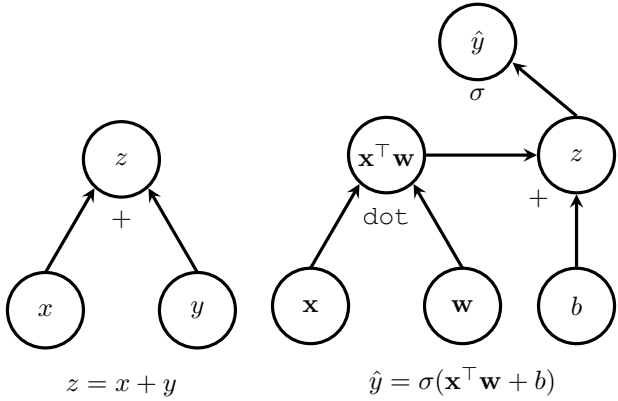

In TensorFlow, machine learning algorithms are represented as
\emph{computational graphs}. A computational or \emph{dataflow} graph is a form
of directed graph where \emph{vertices} or \emph{nodes} describe operations,
while \emph{edges} represent data flowing between these operations. If an output
variable $z$ is the result of applying a binary operation to two inputs $x$ and
$y$, then we draw directed edges from $x$ and $y$ to an output node representing
$z$ and annotate the vertex with a label describing the performed
computation. Examples for computational graphs are given in Figure
\ref{fig:graphs}. The following paragraphs discuss the principle elements of
such a dataflow graph, namely \emph{operations}, \emph{tensors},
\emph{variables} and \emph{sessions}.

\subsubsection{Operations}\label{sec:model-graphs-ops}

The major benefit of representing an algorithm in form of a graph is not only
the intuitive (visual) expression of dependencies between units of a
computational model, but also the fact that the definition of a \emph{node}
within the graph can be kept very general. In TensorFlow, nodes represent
\emph{operations}, which in turn express the combination or transformation of
data flowing through the graph \cite{tensorflow}. An operation can have
\emph{zero or more} inputs and produce \emph{zero or more} outputs. As such, an
operation may represent a mathematical equation, a variable or constant, a
control flow directive, a file I/O operation or even a network communication
port. It may seem unintuitive that an operation, which the reader may associate
with a \emph{function} in the mathematical sense, can represent a constant or
variable. However, a constant may be thought of as an operation that takes no
inputs and always produces the same output corresponding to the constant it
represents. Analogously, a variable is really just an operation taking no input
and producing the current state or value of that variable. Table \ref{tab:ops}
gives an overview of different kinds of operations that may be declared in a
TensorFlow graph.

\begin{table}[b!]
  \begin{tabular}{ll}
    \textbf{Category} & \textbf{Examples}
    \\ \toprule
    Element-wise operations & \texttt{Add}, \texttt{Mul}, \texttt{Exp}
    \\
    Matrix operations & \texttt{MatMul}, \texttt{MatrixInverse}
    \\
    Value-producing operations & \texttt{Constant}, \texttt{Variable}
    \\
    Neural network units & \texttt{SoftMax}, \texttt{ReLU}, \texttt{Conv2D}
    \\
    Checkpoint operations & \texttt{Save}, \texttt{Restore}
    \\ \bottomrule
    \end{tabular}
    \label{tab:ops}
    \caption{Examples for TensorFlow operations \cite{tensorflow}.}
\end{table}

Any operation must be backed by an associated implementation. In
\cite{tensorflow} such an implementation is referred to as the operation's
\emph{kernel}. A particular kernel is always specifically built for execution on
a certain kind of device, such as a CPU, GPU or other hardware unit.

\subsubsection{Tensors}\label{sec:model-graphs-tensors}

In TensorFlow, edges represent data flowing from one operation to another and
are referred to as \emph{tensors}. A tensor is a multi-dimensional collection of
homogeneous values with a fixed, static type. The number of dimensions of a
tensor is termed its \emph{rank}. A tensor's \emph{shape} is the tuple
describing its size, i.e. the number of components, in each dimension. In the
mathematical sense, a tensor is the generalization of two-dimensional matrices,
one-dimensional vectors and also scalars, which are simply tensors of rank zero.

In terms of the computational graph, a tensor can be seen as a \emph{symbolic
  handle} to one of the outputs of an operation. A tensor itself does not hold
or store values in memory, but provides only an interface for retrieving the
value referenced by the tensor. When creating an operation in the TensorFlow
programming environment, such as for the expression $x + y$, a tensor object is
returned. This tensor may then be supplied as input to other computations,
thereby connecting the source and destination operations with an edge. By these
means, data flows through a TensorFlow graph.

Next to regular tensors, TensorFlow also provides a \texttt{SparseTensor}
data structure, allowing for a more space-efficient dictionary-like
representation of \emph{sparse tensors} with only few non-zeros entries.

\subsubsection{Variables}\label{sec:model-graphs-vars}

In a typical situation, such as when performing stochastic gradient descent
(SGD), the graph of a machine learning model is executed from start to end
multiple times for a single experiment. Between two such invocations, the
majority of tensors in the graph are destroyed and do not persist. However, it
is often necessary to maintain state across evaluations of the graph, such as
for the weights and parameters of a neural network. For this purpose, there
exist \emph{variables} in TensorFlow, which are simply special operations that
can be added to the computational graph.

In detail, variables can be described as persistent, mutable handles to
in-memory buffers storing tensors. As such, variables are characterized by a
certain shape and a fixed type. To manipulate and update variables, TensorFlow
provides the \texttt{assign} family of graph operations.

When creating a variable node for a TensorFlow graph, it is necessary to supply
a tensor with which the variable is initialized upon graph execution. The shape
and data type of the variable is then deduced from this
initializer. Interestingly, the variable itself does not store this initial
tensor. Rather, constructing a variable results in the addition of \emph{three}
distinct nodes to the graph:

\begin{enumerate}
  \item The actual variable node, holding the mutable state.
  \item An operation producing the initial value, often a constant.
  \item An \emph{initializer} operation, that \texttt{assign}s the initial value
    to the variable tensor upon evaluation of the graph.
\end{enumerate}

\begin{figure}
  \centering
    \begin{tikzpicture}
    \draw [very thick] (1, 2) circle [radius=0.5cm] node {$v'$};

    \draw [very thick] (0, 0) circle [radius=0.5cm] node {$v$};
    \draw [very thick] (2, 0) circle [radius=0.5cm] node {$i$};

    \draw [very thick, -stealth] (0, 0.5) -- ++(60:1.29);
    \draw [very thick, -stealth] (2, 0.5) -- ++(120:1.29);

    \draw (2.5, 2) node {\texttt{assign}};

    \draw (1, -1) node {$v = i$};
  \end{tikzpicture}
  \caption{The three nodes that are added to the computational graph for every
    variable definition. The first, $v$, is the variable operation that holds a
    mutable in-memory buffer containing the value tensor of the variable. The
    second, $i$, is the node producing the initial value for the variable, which
    can be any tensor. Lastly, the \texttt{assign} node will set the variable to
    the initializer's value when executed. The \texttt{assign} node also
    produces a tensor referencing the initialized value $v'$ of the variable,
    such that it may be connected to other nodes as necessary (e.g. when using a
    variable as the initializer for another variable).}
  \label{fig:variable}
\end{figure}
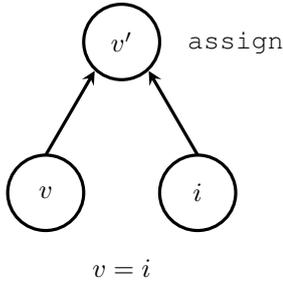

An example for this is given in Figure \ref{fig:variable}.

\subsubsection{Sessions}\label{sec:model-graphs-sessions}

In TensorFlow, the execution of operations and evaluation of tensors may only be
performed in a special environment referred to as \emph{session}. One of the
responsibilities of a session is to encapsulate the allocation and management of
resources such as variable buffers. Moreover, the \texttt{Session} interface of
the TensorFlow library provides a \texttt{run} routine, which is the primary
entry point for executing parts or the entirety of a computational graph. This
method takes as input the nodes in the graph whose tensors should be computed
and returned. Moreover, an optional mapping from arbitrary nodes in the graph to
respective replacement values --- referred to as \emph{feed nodes} --- may be
supplied to \texttt{run} as well \cite{tensorflow}.

Upon invocation of \texttt{run}, TensorFlow will start at the requested output
nodes and work backwards, examining the graph dependencies and computing the
full transitive closure of all nodes that must be executed. These nodes may then
be assigned to one or many physical execution units (CPUs, GPUs etc.) on one or
many machines. The rules by which this assignment takes place are determined by
TensorFlow's \emph{placement algorithm}, discussed in detail in Subsection
\ref{subsec:model-exec}. Furthermore, as there exists the possibility to specify
explicit orderings of node evaluations, called \emph{control dependencies}, the
execution algorithm will ensure that these dependencies are maintained.

\subsection{Execution Model}\label{sec:model-exec}

To execute computational graphs composed of the various elements just discussed,
TensorFlow divides the tasks for its implementation among four distinct groups:
the \emph{client}, the \emph{master}, a set of \emph{workers} and lastly a
number of \emph{devices}. When the client requests evaluation of a TensorFlow
graph via a \texttt{Session}'s \texttt{run} routine, this query is sent to the
master process, which in turn delegates the task to one or more worker processes
and coordinates their execution. Each worker is subsequently responsible for
overseeing one or more devices, which are the physical processing units for
which the kernels of an operation are implemented.

\begin{figure}
  \begin{tikzpicture}
    \draw [very thick] (0, 0) rectangle ++(1.25, 0.75) node [midway] {client};

    \draw [very thick] (2.5, 0) rectangle ++(1.25, 0.75) node [midway] {master};
    \draw [very thick, -stealth] (1.25, 0.375) -- (2.5, 0.375) node [midway,
    above] {\texttt{run}};

    \draw [very thick] (5.25, 0.55) rectangle ++(2.75, 1.5);
    \draw (6.75, 1.8) node {worker $A$};
    \draw (5.45, 0.95) rectangle ++(1.1, 0.5) node [midway] {\small$GPU_0$};
    \draw (6.75, 0.95) rectangle ++(1.1, 0.5) node [midway] {\small$CPU_0$};
    \draw (6.65, 0.75) node {\large\dots};

    \draw [very thick] (5.25, -1.3) rectangle ++(2.75, 1.5);
    \draw (6.75, -0.05) node {worker $B$};
    \draw (5.45, -0.9) rectangle ++(1.1, 0.5) node [midway] {\small$CPU_0$};
    \draw (6.75, -0.9) rectangle ++(1.1, 0.5) node [midway] {\small$CPU_1$};
    \draw (6.65, -1.1) node {\large\dots};

    \draw [very thick, -stealth]
          (3.75, 0.375) .. controls +(1, 0) and (4.2, 1.45) .. (5.25, 1.4);
    \draw [very thick, -stealth]
          (3.75, 0.375) .. controls +(1, 0) and (4.2, -0.55) .. (5.25, -0.5);

  \end{tikzpicture}
  \caption{A visualization of the different execution agents in a multi-machine,
    multi-device hardware configuration.}
  \label{fig:exec}
\end{figure}
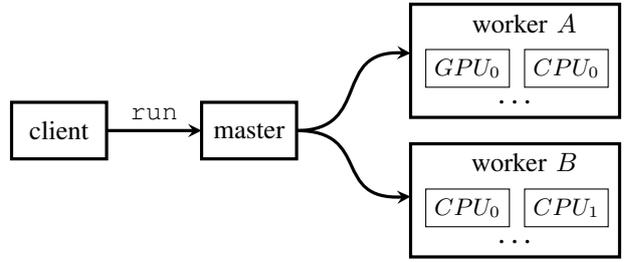

Within this model, there are two degrees of scalability. The first degree
pertains to scaling the number of machines on which a graph is executed. The
second degree refers to the fact that on each machine, there may then be more
than one device, such as, for example, five independent GPUs and/or three
CPUs. For this reason, there exist two ``versions'' of TensorFlow, one for local
execution on a single machine (but possibly many devices), and one supporting a
\emph{distributed} implementation across many machines and many devices. Figure
\ref{fig:exec} visualizes a possible distributed setup. While the initial
release of TensorFlow supported only single-machine execution, the distributed
version was open-sourced on April 13, 2016 \cite{tensorflowdist}.

\subsubsection{Devices}\label{sec:model-exec-devices}

Devices are the smallest, most basic entities in the TensorFlow execution
model. All nodes in the graph, that is, the kernel of each operation, must
eventually be mapped to an available device to be executed. In practice, a
device will most often be either a CPU or a GPU. However, TensorFlow supports
registration of further kinds of physical execution units by the user. For
example, in May 2016, Google announced its \emph{Tensor Processing Unit} (TPU),
a custom built ASIC (application-specific-integrated-circuit) optimized
specifically for fast tensor computations \cite{tpu}. It is thus understandably
easy to integrate new device classes as novel hardware emerges.

To oversee the evaluation of nodes on a device, a worker process is spawned by
the master. As a worker process may manage one or many devices on a single
machine, a device is identified not only by a name, but also an index for its
worker group. For example, the first CPU in a particular group may be identified
by the string ``/cpu:0''.

\subsubsection{Placement Algorithm}\label{sec:model-exec-placement}

To determine what nodes to assign to which device, TensorFlow makes use of a
\emph{placement algorithm}. The placement algorithm simulates the execution of
the computational graph and traverses its nodes from input tensors to output
tensors. To decide on which of the available devices
$\mathbb{D} = \{d_1, \dots, d_n\}$ to place a given node $\nu$ encountered
during this traversal, the algorithm consults a \emph{cost model}
$C_\nu(d)$. This cost model takes into account four pieces of information to
determine the optimal device $\hat{d} = \argmin_{d \in \mathbb{D}} C_\nu(d)$ on
which to place the node during execution:

\begin{enumerate}
  \item Whether or not there exists an implementation (kernel) for a node on the
    given device at all. For example, if there is no GPU kernel for a particular
    operation, any GPU device would automatically incur an infinite cost.
  \item Estimates of the size (in bytes) for a node's input and output tensors.
  \item The expected execution time for the kernel on the device.
  \item A heuristic for the cost of cross-device (and possibly cross-machine)
    transmission of the input tensors to the operation, in the case that the
    input tensors have been placed on nodes different from the one currently
    under consideration.
\end{enumerate}

\subsubsection{Cross-Device Execution}\label{sec:model-exec-single}

\begin{figure}
  \centering
  \begin{subfigure}[b]{0.30\textwidth}
    \centering
    \begin{tikzpicture}
      \draw [very thick] (0, 0) rectangle (2, 1);
      \draw (1, -0.4) node {Device $A$};
      \draw [thick] (1, 0.5) circle [radius=0.25] node {$\nu$};

      \draw [very thick] (4, -0.5) rectangle (6.5, 1.5);
      \draw (5.25, -0.9) node {Device $B$};
      \draw [thick] (5.25, 0.05) circle [radius=0.25] node {$\beta$};
      \draw [thick] (5.25, 0.95) circle [radius=0.25] node {$\alpha$};

      \draw [very thick, -stealth] (1.25, 0.5)
            .. controls (2.5, 0.5) and (3, 1.1)
            ..         (5, 0.95);
      \draw [very thick, -stealth] (1.25, 0.5)
            .. controls (2.5, 0.5) and (3, -0.1)
            ..          (5, 0.05);
    \end{tikzpicture}
    \caption{}
    \label{fig:cross-a}
  \end{subfigure}
  \begin{subfigure}[b]{0.30\textwidth}
    \begin{tikzpicture}
      \draw [very thick] (0, 0) rectangle (2, 1);
      \draw (1, -0.4) node {Device $A$};
      \draw [thick] (0.4, 0.5) circle [radius=0.25] node {$\nu$};

      \draw [very thick] (4, -0.5) rectangle (6.5, 1.5);
      \draw (5.25, -0.9) node {Device $B$};
      \draw [thick] (5.9, 0.05) circle [radius=0.25] node {$\beta$};
      \draw [thick] (5.9, 0.95) circle [radius=0.25] node {$\alpha$};

      \draw [thick] (1.1, 0.3) rectangle ++(0.75, 0.4) node [midway]
      {\footnotesize\texttt{send}};
      \draw [very thick, -stealth] (0.7, 0.5) -- (1.1, 0.5);

      \draw [thick] (4.4, 0.75) rectangle ++(0.75, 0.4) node [midway]
      {\footnotesize\texttt{recv}};
      \draw [very thick, -stealth] (5.15, 0.95) -- ++(0.5, 0);
      \draw [thick] (4.4, -0.15) rectangle ++(0.75, 0.4) node [midway]
      {\footnotesize\texttt{recv}};
      \draw [very thick, -stealth] (5.15, 0.05) -- ++(0.5, 0);

      \draw [very thick, -stealth] (1.85, 0.5)
            .. controls (2.5, 0.5) and (3, 1.1)
            ..         (4.38, 0.95);
      \draw [very thick, -stealth] (1.85, 0.5)
            .. controls (2.5, 0.5) and (3, -0.1)
            ..          (4.38, 0.05);
    \end{tikzpicture}
    \caption{}
    \label{fig:cross-b}
  \end{subfigure}
  \begin{subfigure}[h]{0.30\textwidth}
    \begin{tikzpicture}
      \draw [very thick] (0, 0) rectangle (2, 1);
      \draw (1, -0.4) node {Device $A$};
      \draw [thick] (0.4, 0.5) circle [radius=0.25] node {$\nu$};

      \draw [very thick] (4, -0.5) rectangle (6.5, 1.5);
      \draw (5.25, -0.9) node {Device $B$};
      \draw [thick] (5.9, 0.05) circle [radius=0.25] node {$\beta$};
      \draw [thick] (5.9, 0.95) circle [radius=0.25] node {$\alpha$};

      \draw [thick] (1.1, 0.3) rectangle ++(0.75, 0.4) node [midway]
      {\footnotesize\texttt{send}};
      \draw [very thick, -stealth] (0.7, 0.5) -- (1.1, 0.5);

      \draw [thick] (4.4, 0.3) rectangle ++(0.75, 0.4) node [midway]
      {\footnotesize\texttt{recv}};
      \draw [very thick, -stealth] (5.15, 0.5)
                      .. controls +(0.25, 0)
                               .. +(0.65, 0.25);
      \draw [very thick, -stealth] (5.15, 0.5)
                      .. controls +(0.25, 0)
                               .. +(0.65, -0.25);

      \draw [very thick, -stealth] (1.85, 0.475) -- (4.4, 0.475);
    \end{tikzpicture}
    \caption{}
    \label{fig:cross-c}
  \end{subfigure}
  \caption{The three stages of cross-device communication between graph nodes in
    TensorFlow. Figure \ref{fig:cross-a} shows the initial, conceptual
    connections between nodes on different devices. Figure \ref{fig:cross-b}
    gives a more practical overview of how data is actually transmitted across
    devices using \texttt{send} and \texttt{recv} nodes. Lastly, Figure
    \ref{fig:cross-c} shows the final, canonicalized setup, where there is at
    most one \texttt{recv} node per destination device.}
  \label{fig:cross}
\end{figure}
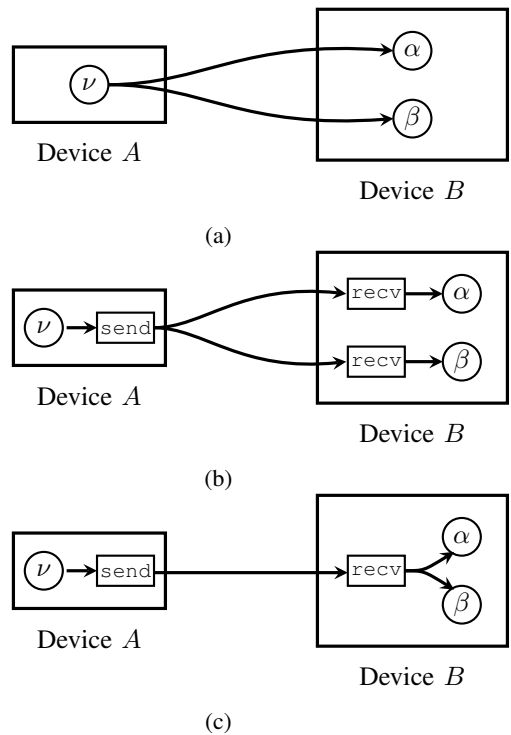

If the hardware configuration of the user's system provides more than one
device, the placement algorithm will often distribute a graph's nodes among
these devices. This can be seen as partitioning the set of nodes into classes,
one per device. As a consequence, there may be cross-device dependencies between
nodes that must be handled via a number of additional steps. Let us consider for
this two devices $A$ and $B$ with particular focus on a node $\nu$ on device
$A$. If $\nu$'s output tensor forms the input to some other operations
$\alpha, \beta$ on device $B$, there initially exist cross-device edges
$\nu \rightarrow \alpha$ and $\nu \rightarrow \beta$ from device $A$ to device
$B$. This is visualized in Figure \ref{fig:cross-a}.

In practice, there must be some means of transmitting $\nu$'s output tensor from
$A$, say a GPU device, to $B$ --- maybe a CPU device. For this reason,
TensorFlow initially replaces the two edges $\nu \rightarrow \alpha$ and
$\nu \rightarrow \beta$ by three new nodes. On device $A$, a \texttt{send} node
is placed and connected to $\nu$. In tandem, on device $B$, two \texttt{recv}
nodes are instantiated and attached to $\alpha$ and $\beta$, respectively. The
\texttt{send} and \texttt{recv} nodes are then connected by two additional
edges. This step is shown in Figure \ref{fig:cross-b}. During execution of the
graph, cross-device communication of data occurs exclusively via these special
nodes. When the devices are located on separate machines, transmission between
the worker processes on these machines may involve remote communication
protocols such as TCP or RDMA.

Finally, an important optimization made by TensorFlow at this step is
``canonicalization'' of $(\mathtt{send}, \mathtt{receive})$ pairs. In the setup
displayed in Figure \ref{fig:cross-b}, the existence of each \texttt{recv} node
on device $B$ would imply allocation and management of a separate buffer to
store $\nu$'s output tensor, so that it may then be fed to nodes $\alpha$ and
$\beta$, respectively. However, an equivalent and more efficient transformation
places only one \texttt{recv} node on device $B$, streams all output from $\nu$
to this single node, and then to the two dependent nodes $\alpha$ and
$\beta$. This last and final evolution is given in Figure \ref{fig:cross-c}.

\subsection{Optimizations}\label{sec:model-optim}

To ensure a maximum of efficiency and performance of the TensorFlow execution
model, a number of optimizations are built into the library. In this subsection,
we examine three such improvements: common subgraph elimination, execution
scheduling and finally lossy compression.

\subsubsection{Common Subgraph Elimination}\label{sec:model-optim-common}

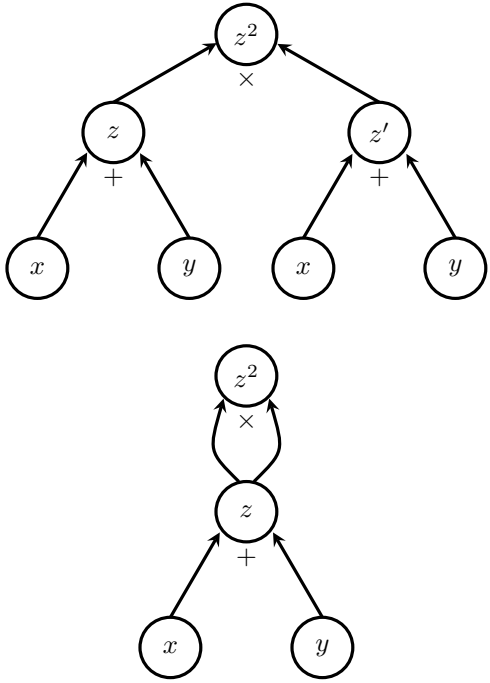
\begin{figure}
  \centering
  \begin{subfigure}[h]{0.5\textwidth}
    \centering
    \begin{tikzpicture}
      \draw [very thick] (1, 1.8) circle [radius=0.4cm] node {$z$};

      \draw [very thick] (0, 0) circle [radius=0.4cm] node {$x$};
      \draw [very thick] (2, 0) circle [radius=0.4cm] node {$y$};

      \draw [very thick, -stealth] (0, 0.4) -- ++(60:1.31);
      \draw [very thick, -stealth] (2, 0.4) -- ++(120:1.3);

      \draw (1, 1.2) node {$+$};

      \draw [very thick] (4.5, 1.8) circle [radius=0.4cm] node {$z'$};

      \draw [very thick] (3.5, 0) circle [radius=0.4cm] node {$x$};
      \draw [very thick] (5.5, 0) circle [radius=0.4cm] node {$y$};

      \draw [very thick, -stealth] (3.5, 0.4) -- ++(60:1.29);
      \draw [very thick, -stealth] (5.5, 0.4) -- ++(120:1.29);

      \draw (4.5, 1.2) node {$+$};

      \draw [very thick] (2.75, 3.1) circle [radius=0.4cm] node {$z^2$};

      \draw [very thick, -stealth] (1, 2.2) -- ++(30:1.57);
      \draw [very thick, -stealth] (4.5, 2.2) -- ++(150:1.55);

      \draw (2.75, 2.5) node {$\times$};

      \draw (0, -1);
    \end{tikzpicture}
  \end{subfigure}
  \begin{subfigure}[h]{0.5\textwidth}
    \centering
    \begin{tikzpicture}
      \draw [very thick] (1, 1.8) circle [radius=0.4cm] node {$z$};

      \draw [very thick] (0, 0) circle [radius=0.4cm] node {$x$};
      \draw [very thick] (2, 0) circle [radius=0.4cm] node {$y$};

      \draw [very thick, -stealth] (0, 0.4) -- ++(60:1.3);
      \draw [very thick, -stealth] (2, 0.4) -- ++(120:1.3);

      \draw (1, 1.2) node {$+$};

      \draw [very thick] (1, 3.6) circle [radius=0.4cm] node {$z^2$};

      \draw (1, 3) node {$\times$};

      \draw [very thick, -stealth] (0.9, 2.2) .. controls +(-0.4, 0.4) .. +(-0.2, 1.1);
      \draw [very thick, -stealth] (1.1, 2.2) .. controls +(+0.4, 0.4) .. +(+0.2, 1.1);
    \end{tikzpicture}
  \end{subfigure}
  \caption{An example of how common subgraph elimination is used to transform the
    equations $z = x + y$, $z' = x + y$, $z^2 = z \cdot z'$ to just two
    equations $z = x + y$ and $z^2 = z \cdot z$. This computation could
    theoretically be optimized further to a \texttt{square} operation requiring
    only one input (thus reducing the cost of data movement), though it is not
    known if TensorFlow employs such secondary canonicalization.}
  \label{fig:subgraph-elim}
\end{figure}

An optimization performed by many modern compilers is \emph{common subexpression
  elimination}, whereby a compiler may possibly replace the computation of an
identical value two or more times by a single instance of that computation. The
result is then stored in a temporary variable and reused where it was previously
re-calculated. Similarly, in a TensorFlow graph, it may occur that the same
operation is performed on identical inputs more than once. This can be
inefficient if the computation happens to be an expensive one. Moreover, it may
incur a large memory overhead given that the result of that operation must be
held in memory multiple times. Therefore, TensorFlow also employs a common
subexpression, or, more aptly put, common \emph{subgraph} elimination pass prior
to execution. For this, the computational graph is traversed and every time two
or more operations of the same type (e.g. \texttt{MatMul}) receiving the same
input tensors are encountered, they are canonicalized to only one such
subgraph. The output tensor of that single operation is then redirected to all
dependent nodes. Figure \ref{fig:subgraph-elim} gives an example of common
subgraph elimination.

\subsubsection{Scheduling}\label{sec:model-optim-schedule}

A simple yet powerful optimization is to schedule node execution as late as
possible. Ensuring that the results of operations remain in memory only for the
minimum required amount of time reduces peak memory consumption and can thus
greatly improve the overall performance of the system. The authors of
\cite{tensorflow} note that this is especially vital on devices such as GPUs,
where memory resources are scarce. Furthermore, careful scheduling also pertains
to the activation of \texttt{send} and \texttt{recv} nodes, where not only
memory but also network resources are contested.

\subsubsection{Lossy Compression}\label{sec:model-optim-lossy}

One of the primary goals of many machine learning algorithms used for
classification, recognition or other tasks is to build robust models. With
\emph{robust} we mean that an optimally trained model should ideally not change
its response if it is first fed a signal and then a noisy variation of that
signal. As such, these machine learning algorithms typically do not require high
precision arithmetic as provided by standard IEEE 754 32-bit floating point
values. Rather, 16 bits of precision in the mantissa would do just as well. For
this reason, another optimization performed by TensorFlow is the internal
addition of conversion nodes to the computational graph, which convert such
high-precision 32-bit floating-point values to truncated 16-bit representations
when communicating across devices and across machines. On the receiving end, the
truncated representation is converted back to 32 bits simply by filling in
zeros, rather than rounding \cite{tensorflow}.

\subsection{Additions to the Basic Programming Model}\label{sec:model-ext}

Having discussed the basic computational paradigms and execution model of
TensorFlow, we will now review three more advanced topics that we deem highly
relevant for anyone wishing to use TensorFlow to create machine learning
algorithms. First, we discuss how TensorFlow handles \emph{gradient
  back-propagation}, an essential concept for many deep learning
applications. Then, we study how TensorFlow graphs support \emph{control
  flow}. Lastly, we briefly touch upon the topic of \emph{checkpoints}, as they
are very useful for maintenance of large models.

\subsubsection{Back-Propagation Nodes}\label{sec:model-ext-backprop}

In a large number of deep learning and other machine learning algorithms, it is
necessary to compute the gradients of particular nodes of the computational
graph with respect to one or many other nodes. For example, in a neural network,
we may compute the cost $c$ of the model for a given example $x$ by passing that
example through a series of non-linear transformations. If the neural network
consists of two hidden layers represented by functions $f(x;w) = f_x(w)$ and
$g(x;w) = g_x(w)$ with internal weights $w$, we can express the cost for that
example as $c = (f_x \circ g_x)(w) = f_x(g_x(w))$. We would then typically
calculate the gradient $dc/dw$ of that cost with respect to the weights and use
it to update $w$. Often, this is done by means of the \emph{back-propagation}
algorithm, which traverses the graph in reverse to compute the chain rule
$[f_x(g_x(w))]' = f_x'(g_x(w)) \cdot g_x'(w)$.

In \cite{goodfellow2016}, two approaches for back-propagating gradients through
a computational graph are described. The first, which the authors refer to as
\emph{symbol-to-number differentiation}, receives a set of input values and then
computes the \emph{numerical values} of the gradients at those input values. It
does so by explicitly traversing the graph first in the forward order
(forward-propagation) to compute the cost, then in reverse order
(back-propagation) to compute the gradients via the chain rule. Another
approach, more relevant to TensorFlow, is what \cite{goodfellow2016} calls
\emph{symbol-to-symbol derivatives} and \cite{tensorflow} terms \emph{automatic
  gradient computation}. In this case, gradients are not computed by an explicit
implementation of the back-propagation algorithm. Rather, special nodes are added
to the computational graph that calculate the gradient of each operation and
thus ultimately the chain rule. To perform back-propagation, these nodes must
then simply be executed like any other nodes by the graph evaluation engine. As
such, this approach does not produce the desired derivatives as a numeric value,
but only as a \emph{symbolic handle} to compute those values.

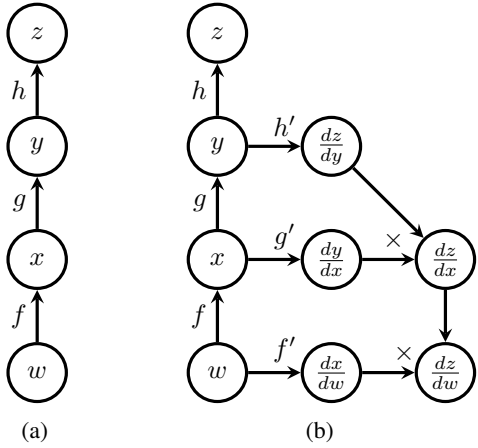
\begin{figure}
  \centering
  \begin{subfigure}[b]{0.2\textwidth}
    \centering
    \begin{tikzpicture}
      \path [very thick] (0, 0)
            coordinate [draw, circle, text width=0.5cm] (w) node {$w$};
      \path [very thick] (0, 1.5)
             coordinate [draw, circle, text width=0.5cm] (x) node {$x$};
      \path [very thick] (0, 3)
            coordinate [draw, circle, text width=0.5cm] (y) node {$y$};
      \path [very thick] (0, 4.5)
            coordinate [draw, circle, text width=0.5cm] (z) node {$z$};

      \draw [very thick, -stealth] (w) -- (x) node [midway, left] {$f$};
      \draw [very thick, -stealth] (x) -- (y) node [midway, left] {$g$};
      \draw [very thick, -stealth] (y) -- (z) node [midway, left] {$h$};
    \end{tikzpicture}
    \caption{}
    \label{fig:gradients-a}
  \end{subfigure}
  \begin{subfigure}[b]{0.2\textwidth}
    \centering
    \begin{tikzpicture}
      \path [very thick] (0, 0)
            coordinate [draw, circle, text width=0.5cm] (w) node {$w$};
      \path [very thick] (0, 1.5)
             coordinate [draw, circle, text width=0.5cm] (x) node {$x$};
      \path [very thick] (0, 3)
            coordinate [draw, circle, text width=0.5cm] (y) node {$y$};
      \path [very thick] (0, 4.5)
            coordinate [draw, circle, text width=0.5cm] (z) node {$z$};

      \draw [very thick, -stealth] (w) -- (x) node [midway, left] {$f$};
      \draw [very thick, -stealth] (x) -- (y) node [midway, left] {$g$};
      \draw [very thick, -stealth] (y) -- (z) node [midway, left] {$h$};

      \path [very thick] (1.5, 0)
            coordinate [draw, circle, text width=0.5cm] (wp) node {$\frac{dx}{dw}$};
      \path [very thick] (1.5, 1.5)
             coordinate [draw, circle, text width=0.5cm] (xp) node {$\frac{dy}{dx}$};
      \path [very thick] (1.5, 3)
            coordinate [draw, circle, text width=0.5cm] (yp) node {$\frac{dz}{dy}$};

      \draw [very thick, -stealth] (w) -- (wp) node [pos=0.7, above] {$f'$};
      \draw [very thick, -stealth] (x) -- (xp) node [pos=0.7, above] {$g'$};
      \draw [very thick, -stealth] (y) -- (yp) node [pos=0.7, above] {$h'$};

      \path [very thick] (3, 0)
            coordinate [draw, circle, text width=0.5cm]
            (dzdw) node {$\frac{dz}{dw}$};
      \path [very thick] (3, 1.5)
             coordinate [draw, circle, text width=0.5cm]
             (dzdx) node {$\frac{dz}{dx}$};

      \draw [very thick, -stealth] (wp) -- (dzdw)
            node [pos=0.8, above] {$\times$};
      \draw [very thick, -stealth] (xp) -- (dzdx)
            node [pos=0.6, above] {$\times$};
      \draw [very thick, -stealth] (yp) -- (dzdx);
      \draw [very thick, -stealth] (dzdx) -- (dzdw);
    \end{tikzpicture}
    \caption{}
    \label{fig:gradients-b}
  \end{subfigure}
  \caption{A computational graph before (\ref{fig:gradients-a}) and after
    (\ref{fig:gradients-b}) gradient nodes are added. In this
    \emph{symbol-to-symbol} approach, the gradient $\frac{dz}{dw}$ is just
    simply an operation like any other and therefore requires no special
    handling by the graph evaluation engine.}
  \label{fig:gradients}
\end{figure}

When TensorFlow needs to compute the gradient of a particular node $\nu$ with
respect to some other tensor $\alpha$, it traverses the graph in reverse order
from $\nu$ to $\alpha$. Each operation $o$ encountered during this traversal
represents a function depending on $\alpha$ and is one of the ``links'' in the
chain $(\nu \,\circ\, \dots\, \circ\, o \,\circ\, \dots)(\alpha)$ producing the
output tensor of the graph. Therefore, TensorFlow adds a \emph{gradient node}
for each such operation $o$ that takes the gradient of the previous link (the
outer function) and multiplies it with its own gradient. At the end of the
traversal, there will be a node providing a symbolic handle to the overall
target derivative $\frac{d\nu}{d\alpha}$, which \emph{implicitly} implements the
back-propagation algorithm. It should now be clear that back-propagation in this
symbol-to-symbol approach is \emph{just another operation}, requiring no
exceptional handling. Figure \ref{fig:gradients} shows how a computational graph
may look before and after gradient nodes are added.

In \cite{tensorflow} it is noted that symbol-to-symbol derivatives may incur a
considerable performance cost and especially result in increased memory
overhead. To see why, it is important to understand that there exist two
equivalent formulations of the chain rule. The first reuses previous
computations and therefore requires them to be stored longer than strictly
necessary for forward-propagation. For arbitrary functions $f$, $g$ and $h$ it is
given in Equation \ref{eq:chain-reuse}:
\begin{equation}\label{eq:chain-reuse}
  \frac{\mathrm{d} f}{\mathrm{d} w} = f'(y) \cdot g'(x) \cdot h'(w) \text{ with } y = g(x), x =
  h(w)
\end{equation}
The second possibility for computing the chain rule was already shown, where
each function recomputes all of its arguments and invokes every function it
depends on. It is given in Equation \ref{eq:chain-recomp} for reference:
\begin{equation}\label{eq:chain-recomp}
  \frac{\mathrm{d} f}{\mathrm{d} w} = f'(g(h(w))) \cdot g'(h(w)) \cdot h'(w)
\end{equation}
According to \cite{tensorflow}, TensorFlow currently employs the first
approach. Given that the inner-most functions must be recomputed for almost
every link of the chain if this approach is not employed, and taking into
consideration that this chain may consist of many hundreds or thousands of
operations, this choice seems sensible. However, on the flip side, keeping
tensors in memory for long periods of time is also not optimal, especially on
devices like GPUs where memory resources are scarce. For Equation
\ref{eq:chain-recomp}, memory held by tensors could in theory be freed as soon
as it has been processed by its graph dependencies. For this reason, in
\cite{tensorflow} the development team of TensorFlow states that recomputing
certain tensors rather than keeping them in memory may be a possible performance
improvement for the future.

\subsubsection{Control Flow}\label{sec:model-ext-flow}

Some machine learning algorithms may benefit from being able to control the flow
of their execution, performing certain steps only under a particular condition
or repeating some computation a fixed or variable number of times. For this,
TensorFlow provides a set of control flow primitives including
\texttt{if}-conditionals and \texttt{while}-loops. The possibility of loops is
the reason why a TensorFlow computational graph is not necessarily
\emph{acyclic}. If the number of iterations for of a loop would be fixed and
known at graph compile-time, its body could be \emph{unrolled} into an acyclic
sequence of computations, one per loop iteration \cite{theano}. However, to
support a variable amount of iterations, TensorFlow is forced to jump through an
additional set of hoops, as described in \cite{tensorflow}.

One aspect that must be especially cared for when introducing control flow is
back-propagation. In the case of a conditional, where an \texttt{if}-operation
returns either one or the other tensor, it must be known which branch was taken
by the node during forward-propagation so that gradient nodes are added only to
this branch. Moreover, when a loop body (which may be a small graph) was
executed a certain number of times, the gradient computation does not only need
to know the number of iterations performed, but also requires access to each
intermediary value produced. This technique of stepping through a loop in
reverse to compute the gradients is referred to as \emph{back-propagation
  through time} in \cite{theano}.

\subsubsection{Checkpoints}\label{sec:model-ext-check}

Another extension to TensorFlow's basic programming model is the notion of
\emph{checkpoints}, which allow for persistent storage and recovery of
variables. It is possible to add \texttt{Save} nodes to the computational graph
and connect them to variables whose tensors you wish to serialize. Furthermore,
a variable may be connected to a \texttt{Restore} operation, which deserializes
a stored tensor at a later point. This is especially useful when training a
model over a long period of time to keep track of the model's performance while
reducing the risk of losing any progress made. Also, checkpoints are a vital
element to ensuring fault tolerance in a distributed environment
\cite{tensorflow}.

\section{The TensorFlow Programming Interface}\label{sec:code}

Having conveyed the abstract concepts of TensorFlow's computational model in
Section \ref{sec:model}, we will now concretize those ideas and speak to
TensorFlow's programming interface. We begin with a brief discussion of the
available language interfaces. Then, we provide a more hands-on look at
TensorFlow's Python API by walking through a simple practical example. Lastly,
we give insight into what higher-level abstractions exist for TensorFlow's API,
which are especially beneficial for rapid prototyping of machine learning
models.

\subsection{Interfaces}\label{sec:code-interfaces}

There currently exist two programming interfaces, in C++ and Python, that permit
interaction with the TensorFlow backend. The Python API boasts a very rich
feature set for creation and execution of computational graphs. As of this
writing, the C++ interface (which is really just the core backend
implementation) provides a comparatively much more limited API, allowing only to
execute graphs built with Python and serialized to Google's Protocol
Buffer\footnote{https://developers.google.com/protocol-buffers/} format. While
there is experimental support for also building computational graphs in C++,
this functionality is currently not as extensive as in Python.

It is noteworthy that the Python API integrates very well with
NumPy\footnote{http://www.numpy.org}, a popular open source Python numeric and
scientific programming library. As such, TensorFlow tensors may be interchanged
with NumPy \texttt{ndarrays} in many places.

\subsection{Walkthrough}\label{sec:code-walk}

In the following paragraphs we give a step-by-step walkthrough of a practical,
real-world example of TensorFlow's Python API. We will train a simple
multi-layer perceptron (MLP) with one input and one output layer to classify
handwritten digits in the MNIST\footnote{http://yann.lecun.com/exdb/mnist/}
dataset. In this dataset, the examples are small images of $28 \times 28$ pixels
depicting handwritten digits in $\in \{0, \dots, 9\}$. We receive each such
example as a flattened vector of $784$ gray-scale pixel intensities. The label
for each example is the digit it is supposed to represent.

We begin our walkthrough by importing the TensorFlow library and reading the
MNIST dataset into memory. For this we assume a utility module
\texttt{mnist\_data} with a method \texttt{read} which expects a path to extract
and store the dataset. Moreover, we pass the parameter \texttt{one\_hot=True} to
specify that each label be given to us as a \emph{one-hot-encoded} vector
$(d_1, \dots, d_{10})^\top$ where all but the $i$-th component are set to zero
if an example represents the digit $i$:

\begin{lstlisting}
import tensorflow as tf

# Download and extract the MNIST data set.
# Retrieve the labels as one-hot-encoded vectors.
mnist = mnist_data.read("/tmp/mnist", one_hot=True)
\end{lstlisting}

Next, we create a new computational graph via the \texttt{tf.Graph}
constructor. To add operations to this graph, we must register it as the
\emph{default graph}. The way the TensorFlow API is designed, library routines
that create new operation nodes always attach these to the current default
graph. We register our graph as the default by using it as a Python context
manager in a \texttt{with-as} statement:

\begin{lstlisting}
# Create a new graph
graph = tf.Graph()

# Register the graph as the default one to add nodes
with graph.as_default():
  # Add operations ...
\end{lstlisting}

We are now ready to populate our computational graph with operations. We begin
by adding two \emph{placeholder} nodes \texttt{examples} and
\texttt{labels}. Placeholders are special variables that \emph{must} be replaced
with concrete tensors upon graph execution. That is, they must be supplied in
the \texttt{feed\_dict} argument to \texttt{Session.run()}, mapping tensors to
replacement values. For each such placeholder, we specify a shape and
data type. An interesting feature of TensorFlow at this point is that we may
specify the Python keyword \texttt{None} for the first dimension of each
placeholder shape. This allows us to later on feed a tensor of variable size in
that dimension. For the column size of the \texttt{example} placeholder, we
specify the number of features for each image, meaning the $28 \times 28 = 784$
pixels. The label placeholder should expect 10 columns, corresponding to the
10-dimensional one-hot-encoded vector for each label digit:

\begin{lstlisting}
# Using a 32-bit floating-point data type tf.float32
examples = tf.placeholder(tf.float32, [None, 784])
labels = tf.placeholder(tf.float32, [None, 10])
\end{lstlisting}

Given an example matrix $\mathbf{X} \in \mathbb{R}^{n \times 784}$ containing
$n$ images, the learning task then applies an affine transformation
$\mathbf{X} \cdot \mathbf{W} + \mathbf{b}$, where $\mathbf{W}$ is a \emph{weight}
  matrix $\in \mathbb{R}^{784 \times 10}$ and $\mathbf{b}$ a \emph{bias} vector
$\in \mathbb{R}^{10}$. This yields a new matrix
$\mathbf{Y} \in \mathbb{R}^{n \times 10}$, containing the \emph{scores} or
\emph{logits} of our model for each example and each possible digit. These
scores are more or less arbitrary values and not a probability distribution,
i.e. they need neither be $\in [0, 1]$ nor sum to one. To transform the logits
into a valid probability distribution, giving the likelihood $\Pr[x = i]$ that
the $x$-th example represents the digit $i$, we make use of the softmax
function, given in Equation \ref{eq:softmax}. Our final estimates are thus
calculated by $\softmax(\mathbf{X} \cdot \mathbf{W} + \mathbf{b})$, as shown
below:

\begin{lstlisting}
# Draw random weights for symmetry breaking
weights = tf.Variable(tf.random_uniform([784, 10]))
# Slightly positive initial bias
bias = tf.Variable(tf.constant(0.1, shape=[10]))
# tf.matmul performs the matrix multiplication XW
# Note how the + operator is overloaded for tensors
logits = tf.matmul(examples, weights) + bias
# Applies the operation element-wise on tensors
estimates = tf.nn.softmax(logits)
\end{lstlisting}

\begin{equation}\label{eq:softmax}
  \softmax(\mathbf{x})_i = \frac{\exp(\mathbf{x}_i)}{\sum_j \exp(\mathbf{x}_j)}
\end{equation}

Next, we compute our objective function, producing the error or \emph{loss} of
the model given its current trainable parameters $\mathbf{W}$ and
$\mathbf{b}$. We do this by calculating the \emph{cross entropy}
$H(\mathbf{L}, \mathbf{Y})_i = -\sum_j \mathbf{L}_{i,j} \cdot
\log(\mathbf{Y}_{i,j})$ between the probability distributions of our estimates
$\mathbf{Y}$ and the one-hot-encoded labels $\mathbf{L}$. More precisely, we
consider the mean cross entropy over all examples as the loss:

\begin{lstlisting}
# Computes the cross-entropy and sums the rows
cross_entropy = -tf.reduce_sum(
    labels * tf.log(estimates), [1])
loss = tf.reduce_mean(cross_entropy)
\end{lstlisting}

Now that we have an objective function, we can run (stochastic) gradient descent
to update the weights of our model. For this, TensorFlow provides a
\texttt{GradientDescentOptimizer} class. It is initialized with the
learning rate of the algorithm and provides an operation \texttt{minimize}, to
which we pass our \texttt{loss} tensor. This is the operation we will run
repeatedly in a \texttt{Session} environment to train our model:

\begin{lstlisting}
# We choose a learning rate of 0.5
gdo = tf.train.GradientDescentOptimizer(0.5)
optimizer = gdo.minimize(loss)
\end{lstlisting}

Finally, we can actually train our algorithm. For this, we enter a session
environment using a \texttt{tf.Session} as a context manager. We pass our graph
object to its constructor, so that it knows which graph to manage. To then
execute nodes, we have several options. The most general way is to call
\texttt{Session.run()} and pass a list of tensors we wish to
compute. Alternatively, we may call \texttt{eval()} on tensors and
\texttt{run()} on operations directly. Before evaluating any other node, we must
first ensure that the variables in our graph are initialized. Theoretically, we
could \texttt{run} the \texttt{Variable.initializer} operation for each
variable. However, one most often just uses the
\texttt{tf.initialize\_all\_variables()} utility operation provided by
TensorFlow, which in turn executes the \texttt{initializer} operation for each
\texttt{Variable} in the graph. Then, we can perform a certain number of
iterations of stochastic gradient descent, fetching an example and label
mini-batch from the MNIST dataset each time and feeding it to the \texttt{run}
routine. At the end, our loss will (hopefully) be small:

\begin{lstlisting}
with tf.Session(graph=graph) as session:
    # Execute the operation directly
    tf.initialize_all_variables().run()
    for step in range(1000):
        # Fetch next 100 examples and labels
        x, y = mnist.train.next_batch(100)
        # Ignore the result of the optimizer (None)
        _, loss_value = session.run(
            [optimizer, loss],
            feed_dict={examples: x, labels: y})
        print('Loss at step {0}: {1}'
                .format(step, loss_value))
\end{lstlisting}

The full code listing for this example, along with some additional
implementation to compute an accuracy metric at each time step is given in
Appendix \ref{app:code}.

\subsection{Abstractions}\label{sec:code-abstract}

You may have observed how a relatively large amount of effort was required to
create just a very simple two-layer neural network. Given that \emph{deep}
learning, by implication of its name, makes use of \emph{deep} neural networks
with many hidden layers, it may seem infeasible to each time create weight and
bias variables, perform a matrix multiplication and addition and finally apply
some non-linear activation function. When testing ideas for new deep learning
models, scientists often wish to rapidly prototype networks and quickly exchange
layers. In that case, these many steps may seem very low-level, repetitive and
generally cumbersome. For this reason, there exist a number of open source
libraries that \emph{abstract} these concepts and provide higher-level building
blocks, such as entire layers. We find
\emph{PrettyTensor}\footnote{https://github.com/google/prettytensor},
\emph{TFLearn}\footnote{https://github.com/tflearn/tflearn} and
\emph{Keras}\footnote{http://keras.io} especially noteworthy. The following
paragraphs give a brief overview of the first two abstraction libraries.

\subsubsection{PrettyTensor}\label{sec:code-abstract-prettytensor}

PrettyTensor is developed by Google and provides a high-level interface to the
TensorFlow API via the \emph{Builder} pattern. It allows the user to wrap
TensorFlow operations and tensors into ``pretty'' versions and then quickly
chain any number of layers operating on these tensors. For example, it is
possible to feed an input tensor into a fully connected (``dense'')
neural network layer as we did in Subsection \ref{sec:code-walk} with just a
single line of code. Shown below is an example use of PrettyTensor, where a
standard TensorFlow placeholder is wrapped into a library-compatible object and
then fed through three fully connected layers to finally output a softmax
distribution.

\begin{lstlisting}
examples = tf.placeholder([None, 784], tf.float32)
softmax = (prettytensor.wrap(examples)
              .fully_connected(256, tf.nn.relu)
              .fully_connected(128, tf.sigmoid)
              .fully_connected(64, tf.tanh)
              .softmax(10))
\end{lstlisting}

\subsubsection{TFLearn}\label{sec:code-abstract-prettytensor}

TFLearn is another abstraction library built on top of TensorFlow that provides
high-level building blocks to quickly construct TensorFlow graphs. It has a
highly modular interface and allows for rapid chaining of neural network layers,
regularization functions, optimizers and other elements. Moreover, while
PrettyTensor still relied on the standard \texttt{tf.Session} setup to train and
evaluate a model, TFLearn adds functionality to easily train a model given an
example batch and corresponding labels. As many TFLearn functions, such as those
creating entire layers, return vanilla TensorFlow objects, the library is well
suited to be mixed with existing TensorFlow code. For example, we could replace
the entire setup for the output layer discussed in Subsection
\ref{sec:code-walk} with just a single TFLearn method invocation, leaving the
rest of our code base \emph{untouched}. Furthermore, TFLearn handles everything
related to visualization with TensorBoard, discussed in Section
\ref{sec:visual}, automatically. Shown below is how we can reproduce the full 65
lines of standard TensorFlow code given in Appendix \ref{app:code} with
\emph{less than 10 lines} using TFLearn.

\begin{lstlisting}
import tflearn
import tflearn.datasets.mnist as mnist

X, Y, validX, validY = mnist.load_data(one_hot=True)

# Building our neural network
input_layer = tflearn.input_data(shape=[None, 784])
output_layer = tflearn.fully_connected(input_layer, 10, activation='softmax')

# Optimization
sgd = tflearn.SGD(learning_rate=0.5)
net = tflearn.regression(output_layer, optimizer=sgd)

# Training
model = tflearn.DNN(net)
model.fit(X, Y, validation_set=(validX, validY))
\end{lstlisting}

\section{Visualization of TensorFlow Graphs}\label{sec:visual}

Deep learning models often employ neural networks with a highly complex and
intricate structure. For example, \cite{inception} reports of deep convolutional
network based on the Google \emph{Inception} model with more than 36,000
individual units, while \cite{tensorflow} states that certain long short-term
memory (LSTM) architectures can span over 15,000 nodes. To maintain a clear
overview of such complex networks, facilitate model debugging and allow
inspection of values on various levels of detail, powerful visualization tools
are required. \emph{TensorBoard}, a web interface for graph visualization and
manipulation built directly into TensorFlow, is an example for such a tool. In
this section, we first list a number of noteworthy features of TensorBoard and
then discuss how it is used from TensorFlow's programming interface.

\subsection{TensorBoard Features}\label{sec:visual-features}

\begin{figure}[h!]
  \centering
  \begin{subfigure}[h]{0.5\textwidth}
    \centering
    \includegraphics[scale=0.4]{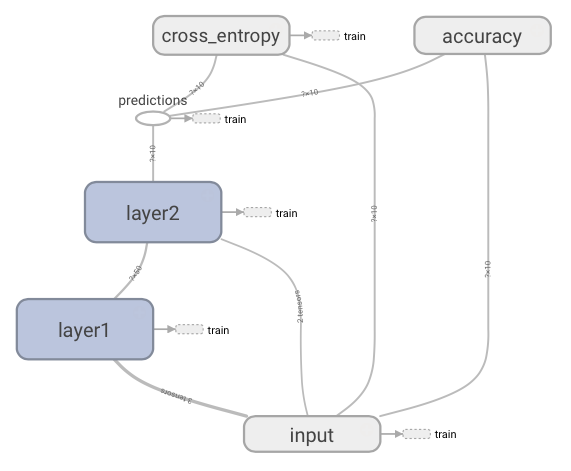}
   \caption{}
   \label{fig:tensorboard-a}
  \end{subfigure}

  \vspace{0.3cm}

  \begin{subfigure}[h]{0.5\textwidth}
    \centering
    \includegraphics[scale=0.4]{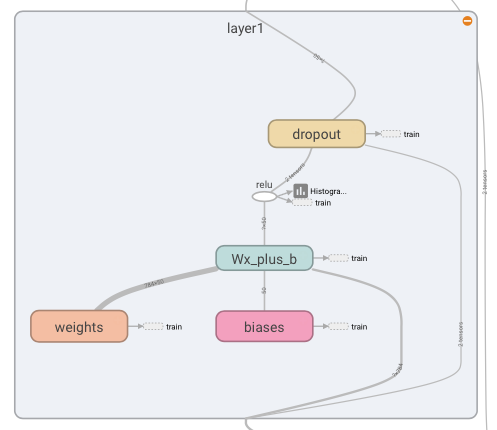}
    \caption{}
    \label{fig:tensorboard-b}
  \end{subfigure}

  \vspace{0.3cm}

  \begin{subfigure}[h]{0.2\textwidth}
    \centering
    \includegraphics[scale=0.35]{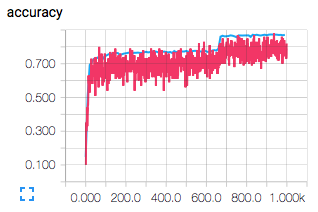}
    \caption{}
    \label{fig:tensorboard-c}
  \end{subfigure}
  \begin{subfigure}[h]{0.2\textwidth}
    \centering
    \includegraphics[scale=0.35]{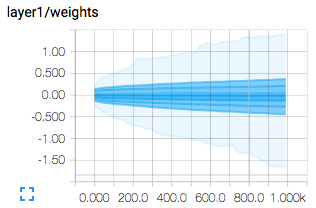}
    \caption{}
    \label{fig:tensorboard-d}
  \end{subfigure}
  \caption{A demonstration of Tensorboard's graph visualization features. Figure
    \ref{fig:tensoroard-a} shows the complete graph, while Figure
    \ref{fig:tensorboard-b} displays the expansion of the first layer. Figures
    \ref{fig:tensorboard-c} and \ref{fig:tensorboard-d} give examples for scalar
    and history summaries, respectively.}
  \label{fig:tensorboard}
\end{figure}

The core feature of TensorBoard is the lucid visualization of computational
graphs, exemplified in Figure \ref{fig:tensorboard-a}. Graphs with complex
topologies and many layers can be displayed in a clear and organized manner,
allowing the user to understand exactly how data flows through it. Especially
useful is TensorBoard's notion of \emph{name scopes}, whereby nodes or entire
subgraphs may be grouped into one visual block, such as a single neural network
layer. Such name scopes can then be expanded interactively to show the grouped
units in more detail. Figure \ref{fig:tensorboard-b} shows the expansion of one
the name scopes of Figure \ref{fig:tensorboard-a}.

Furthermore, TensorBoard allows the user to track the development of individual
tensor values over time. For this, you can attach two kinds of \emph{summary
  operations} to nodes of the computational graph: \emph{scalar summaries} and
\emph{histogram summaries}. Scalar summaries show the progression of a scalar
tensor value, which can be sampled at certain iteration counts. In this way, you
could, for example, observe the accuracy or loss of your model with
time. Histogram summary nodes allow the user to track value
\emph{distributions}, such as those of neural network weights or the final
softmax estimates. Figures \ref{fig:tensorboard-c} and \ref{fig:tensorboard-d}
give examples of scalar and histogram summaries, respectively. Lastly,
TensorBoard also allows visualization of images. This can be useful to show the
images sampled for each mini-batch of an image classification task, or to
visualize the kernel filters of a convolutional neural network
\cite{tensorflow}.

We note especially how interactive the TensorBoard web interface is. Once your
computational graph is uploaded, you can pan and zoom the model as well as
expand or contract individual name scopes. A demonstration of TensorBoard is
available at \emph{https://www.tensorflow.org/tensorboard/index.html}.

\subsection{TensorBoard in Practice}\label{sec:visual-code}

To integrate TensorBoard into your TensorFlow code, three steps are
required. Firstly, it is wise to group nodes into \emph{name scopes}. Then, you
may add scalar and histogram summaries to you operations. Finally, you must
instantiate a \texttt{SummaryWriter} object and hand it the tensors produced by
the summary nodes in a session context whenever you wish to store new
summaries. Rather than fetching individual summaries, it is also possible to
combine all summary nodes into one via the \texttt{tf.merge\_all\_summaries()}
operation.

\begin{lstlisting}
with tf.name_scope('Variables'):
    x = tf.constant(1.0)
    y = tf.constant(2.0)
    tf.scalar_summary('z', x + y)

merged = tf.merge_all_summaries()
writer = tf.train.SummaryWriter('/tmp/log', graph)

with tf.Session(graph=graph):
    for step in range(1000):
        writer.add_summary(
            merged.eval(), global_step=step)
\end{lstlisting}

\section{Comparison With Other Deep Learning Frameworks}\label{sec:comp}

Next to TensorFlow, there exist a number of other open source deep learning
software libraries, the most popular being Theano, Torch and Caffe. In this
section, we explore the \emph{qualitative} as well as \emph{quantitative}
differences between TensorFlow and each of these alternatives. We begin with a
``high level'' qualitative comparison and examine where TensorFlow diverges or
overlaps conceptually or architecturally. Then, we review a few sources of
quantitative comparisons and state as well as discuss their results.

\subsection{Qualitative Comparison}\label{sec:comp-quality}

The following three paragraphs compare Theano, Torch and Caffe to TensorFlow,
respectively. Table \ref{tab:comp} provides an overview of the most important
talking points.

\subsubsection{Theano}\label{sec:comp-quality-theano}

Of the three candidate alternatives we discuss, Theano, which has a Python
frontend, is most similar to TensorFlow. Like TensorFlow, Theano's programming
model is \emph{declarative} rather than \emph{imperative} and based on
computational graphs. Also, Theano employs symbolic differentiation, as does
TensorFlow. However, Theano is known to have very long graph compile times as it
translates Python code to C++/CUDA \cite{theano}. In part, this is due to the
fact that Theano applies a number of more advanced graph optimization algorithms
\cite{theano}, while TensorFlow currently only performs common subgraph
elimination. Moreover, Theano's visualization tools are very poor in comparison
to TensorBoard. Next to built-in functionality to output plain text
representations of the graph or static images, a plugin can be used to generate
slightly interactive HTML visualizations. However, it is nowhere near as
powerful as TensorBoard. Lastly, there is also no (out-of-the-box) support for
distributing the execution of a computational graph, while this is a key feature
of TensorFlow.

\subsubsection{Torch}\label{sec:comp-quality-torch}

One of the principle differences between Torch and TensorFlow is the fact that
Torch, while it has a C/CUDA backend, uses Lua as its main frontend. While
Lua(JIT) is one of the fastest scripting languages and allows for rapid
prototyping and quick execution, it is not yet very mainstream. This implies
that while it may be easy to train and develop models with Torch, Lua's limited
API and library ecosystem can make industrial deployment harder compared to a
Python-based library such as TensorFlow (or Theano). Besides the language
aspect, Torch's programming model is fundamentally quite different from
TensorFlow. Models are expressed in an imperative programming style and not as
declarative computational graphs. This means that the programmer must, in fact,
be concerned with the order of execution of operations. It also implies that
Torch does not use symbol-to-symbol, but rather symbol-to-number differentiation
requiring explicit forward and backward passes to compute gradients.

\subsubsection{Caffe}\label{sec:comp-quality-caffe}

Caffe is most dissimilar to TensorFlow --- in various ways. While there exist
high-level MATLAB and Python frontends to Caffe for model creation, its main
interface is really the Google Protobuf ``language'' (it is more a fancy, typed
version of JSON), which gives a very different experience compared to
Python. Also, the basic building blocks in Caffe are not operations, but entire
neural network layers. In that sense, TensorFlow can be considered fairly
low-level in comparison. Like Torch, Caffe has no notion of a computational
graphs or symbols and thus computes derivatives via the symbol-to-number
approach. Caffe is especially well suited for development of convolutional
neural networks and image recognition tasks, however it falls short in many
other state-of-the-art categories supported well by TensorFlow. For example,
Caffe, by design, does not support cyclic architectures, which form the basis of
RNN, LSTM and other models. Caffe has no support for distributed
execution\footnote{\url{https://github.com/BVLC/caffe/issues/876}}.

\begin{table}[h!]
  \begin{tabular}{llllc}
    \textbf{Library} & \textbf{Frontends} &
    \textbf{Style} &
    \textbf{Gradients} &
    \specialcell{\textbf{Distributed}\\\textbf{Execution}}
    \\ \toprule
    TensorFlow & Python, C++\textsuperscript{\dag} &
    Declarative & Symbolic & $\checkmark$\textsuperscript{\ddag}
    \\
    Theano & Python & Declarative & Symbolic & $\times$
    \\
    Torch & LuaJIT & Imperative & Explicit & $\times$
    \\
    Caffe & Protobuf & Imperative & Explicit & $\times$
    \\ \bottomrule
  \end{tabular}
  \begin{tabular}{l}
    \textsuperscript{\dag} Very limited API.
    \\
    \textsuperscript{\ddag} Starting with TensorFlow 0.8, released in April 2016
    \cite{tensorflowdist}.
  \end{tabular}
  \caption{A table comparing TensorFlow to Theano, Torch and Caffe in several
    categories.}
  \label{tab:comp}
\end{table}

\subsection{Quantitative Comparison}\label{sec:comp-quantity}

We will now review three sources of quantitative comparisons between TensorFlow
and other deep learning libraries, providing a summary of the most important
results of each work. Furthermore, we will briefly discuss the overall trend of
these benchmarks.

The first work, \cite{bosch}, authored by the Bosch Research and Technology
Center in late March 2016, compares the performance of TensorFlow, Torch, Theano
and Caffe (among others) with respect to various neural network architectures.
Their setup involves Ubuntu 14.04 running on an Intel Xeon E5-1650 v2 CPU @ 3.50
GHz and an NVIDIA GeForce GTX Titan X/PCIe/SSE2 GPU. One benchmark we find
noteworthy tests the relative performance of each library on a slightly modified
reproduction of the LeNet CNN model \cite{lenet}. More specifically, the authors
measure the forward-propagation time, which they deem relevant for model
deployment, and the back-propagation time, important for model training. We have
reproduced an excerpt of their results in Table \ref{tab:bosch}, where we show
their outcomes on (a) a CPU running 12 threads and (b) a GPU. Interestingly, for
(a), TensorFlow ranks second behind Torch in both the forward and backward
measure while in (b) TensorFlow's performance drops significantly, placing it
\emph{last} in both categories. The authors of \cite{bosch} note that one reason
for this may be that they used the NVIDIA cuDNN \emph{v2} library for their GPU
implementation with TensorFlow while using cuDNN \emph{v3} for the others. They
state that as of their writing, this was the recommended configuration for
TensorFlow\footnote{As of TensorFlow 0.8, released in April 2016 and thus after
  the publication of \cite{bosch}, TensorFlow now supports cuDNN v4, which
  promises better performance on GPUs than cuDNN v3 and especially cuDNN v2.}.

\begin{table}
  \begin{subfigure}[h]{0.49\textwidth}
    \centering
    \begin{tabular}{ccc}
      \textbf{Library} & \textbf{Forward (ms)} & \textbf{Backward (ms)}
      \\ \toprule
      TensorFlow & 16.4 & 50.1
      \\
      Torch & \textbf{4.6} & \textbf{16.5}
      \\
      Caffe & 33.7 & 66.4
      \\
      Theano & 78.7 & 204.3
      \\ \bottomrule
    \end{tabular}
    \caption{CPU (12 threads)}
  \label{tab:bosch-a}
  \end{subfigure}

  \vspace{0.5cm}

  \begin{subfigure}[h]{0.49\textwidth}
    \centering
    \begin{tabular}{ccc}
      \textbf{Library} & \textbf{Forward (ms)} & \textbf{Backward (ms)}
      \\ \toprule
      TensorFlow & 4.5 & 14.6
      \\
      Torch & \textbf{0.5} & 1.7
      \\
      Caffe & 0.8 & 1.9
      \\
      Theano & \textbf{0.5} & \textbf{1.4}
      \\ \bottomrule
    \end{tabular}
    \caption{GPU}
    \label{tab:bosch-b}
  \end{subfigure}
  \caption{This table shows the benchmarks performed by \cite{bosch}, where
    TensorFlow, Torch, Caffe and Theano are compared on a LeNet model
    reproduction \cite{lenet}. \ref{tab:bosch-a} shows the results performed
    with 12 threads each on a CPU, while \ref{tab:bosch-b} gives the outcomes on
    a graphics chips.}
  \label{tab:bosch}
\end{table}

The second source in our collection is the \emph{convnet-benchmarks} repository
on GitHub by Soumith Chintala \cite{convnet-bench}, an artificial intelligence
research engineer at Facebook. The commit we reference\footnote{Commit sha1
  hash: 84b5bb1785106e89691bc0625674b566a6c02147} is dated April 25,
2016. Chintala provides an extensive set of benchmarks for a variety of
convolutional network models and includes many libraries, including TensorFlow,
Torch and Caffe in his measurements. Theano is not present in all tests, so we
will not review its performance in this benchmark suite. The author's hardware
configuration is a 6-core Intel Core i7-5930K CPU @ 3.50GHz and an NVIDIA Titan
X graphics chip running on Ubuntu 14.04. Inter alia, Chintala gives the forward
and backward-propagation time of TensorFlow, Torch and Caffe for the AlexNet CNN
model \cite{alexnet}. In these benchmarks, TensorFlow performs second-best in
both measures behind Torch, with Caffe lagging relatively far behind. We
reproduce the relevant results in Table \ref{tab:convnet}.

\begin{table}
  \centering
  \begin{tabular}{ccc}
    \textbf{Library} & \textbf{Forward (ms)} & \textbf{Backward (ms)}
    \\ \toprule
    TensorFlow & 26  & 55
    \\
    Torch & \textbf{25} & \textbf{46}
    \\
    Caffe & 121 & 203
    \\
    Theano & \textendash & \textendash
    \\ \bottomrule
  \end{tabular}
  \caption{The result of Soumith Chintala's benchmarks for TensorFlow, Torch and
    Caffe (not Theano) on an AlexNet ConvNet model \cite{alexnet,
      convnet-bench}.}
  \label{tab:convnet}
\end{table}

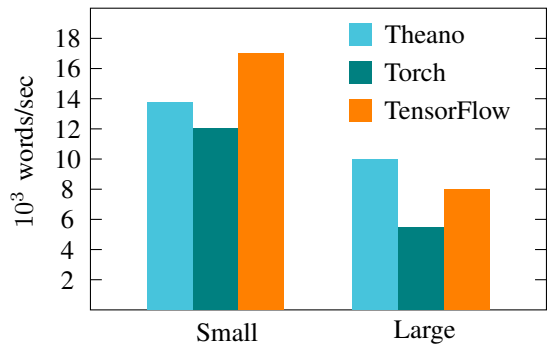
\begin{figure}
  \centering
  \begin{tikzpicture}
    \draw (0, 0) rectangle (6, 4);

    \foreach \i in {2, 4, ..., 18} {
      \draw (0, {\i / 5}) -- ++(+0.15, 0);
      \draw (5.85, {\i / 5}) -- ++(+0.15, 0);
      \draw (-0.3, {\i / 5}) node {\i};
    }

    \draw (-0.9, 2.1) node [rotate=90] {$10^3$ words/sec};

    \fill [SkyBlue] (3.4, 3.5) rectangle ++(0.3, 0.3);
    \draw (4.4, 3.65) node {Theano};

    \fill [teal] (3.4, 3) rectangle ++(0.3, 0.3);
    \draw (4.275, 3.15) node {Torch};

    \fill [orange] (3.4, 2.5) rectangle ++(0.3, 0.3);
    \draw (4.7, 2.65) node {TensorFlow};


    \fill [SkyBlue] (0.75, 0) rectangle ++(0.6, 2.75);
    \fill [teal] (1.35, 0) rectangle ++(0.6, 2.4);
    \fill [orange] (1.95, 0) rectangle ++(0.6, 3.4);

    \draw (1.8, -0.3) node {Small};


    \fill [SkyBlue] (3.45, 0) rectangle ++(0.6, 2);
    \fill [teal] (4.05, 0) rectangle ++(0.6, 1.1);
    \fill [orange] (4.65, 0) rectangle ++(0.6, 1.6);

    \draw (4.4, -0.3) node {Large};

  \end{tikzpicture}
  \caption{The results of \cite{theano}, comparing TensorFlow, Theano and Torch
    on an LSTM model for the Penn Treebank dataset \cite{penntreebank}. On the
    left the authors tested a small model with a single hidden layer and 200
    units; on the right they use two layers with 650 units each. }
  \label{fig:theano-results}
\end{figure}

Lastly, we review the results of \cite{theano}, published by the Theano
development team on May 9, 2016. Next to a set of benchmarks for four popular
CNN models, including the aforementioned AlexNet architecture, the work also
includes results for an LSTM network operating on the Penn Treebank dataset
\cite{penntreebank}. Their benchmarks measure words processed per second for a
small model consisting of a single 200-unit hidden layer with sequence length
20, and a large model with two 650-unit hidden layers and a sequence length of
50. In \cite{theano} also a medium-sized model is tested, which we ignore for
our review. The authors state a hardware configuration consisting of an NVIDIA
Digits DevBox with 4 Titan X GPUs and an Intel Core i7-5930K CPU. Moreover, they
used cuDNN v4 for all libraries included in their benchmarks, which are
TensorFlow, Torch and Theano. Results for Caffe are not given. In their
benchmarks, TensorFlow performs best among all three for the small model,
followed by Theano and then Torch. For the large model, TensorFlow is placed
second behind Theano, while Torch remains in last place. Table
\ref{fig:theano-results} shows these results, taken from \cite{theano}.

When TensorFlow was first released, it performed poor on benchmarks, causing
disappointment within the deep learning community. Since then, new releases of
TensorFlow have emerged, bringing with them improving results. This is reflected
in our selection of works. The earliest of the three sources, \cite{bosch},
published in late March 2016, ranks TensorFlow consistently uncompetitive
compared to Theano, Torch and Caffe. Released almost two months later,
\cite{convnet-bench} ranks TensorFlow comparatively better. The latest work
reviewed, \cite{theano}, then places TensorFlow in first or second place for
LSTMs and also other architectures discussed by the authors. We state that one
reason for this upward trend is that \cite{theano} uses TensorFlow with cuDNN v4
for its GPU experiments, whereas \cite{bosch} still used cuDNN v2. While we
predict that TensorFlow will improve its performance on measurements similar to
the ones discussed in the future, we believe that these benchmarks --- also
today --- do not make full use of TensorFlow's potential. The reason for this is
that all tests were performed on a \emph{single} machine. As we reviewed in
depth in section \ref{sec:model-exec}, TensorFlow was built with massively
parallel distributed computing in mind. This ability is currently unique to
TensorFlow among the popular deep learning libraries and we estimate that it
would be advantageous to its performance, particularly for large-scale
models. We thus hope to see more benchmarks in literature in the future, making
better use of TensorFlow's many-machine, many-device capabilities.

\section{Use Cases of TensorFlow Today}\label{sec:uses}

In this section, we investigate where TensorFlow is already in use today. Given
that TensorFlow was released only little over 6 months ago as of this writing,
its adoption in academia and industry is not yet widespread. Migration from an
existing system based on some other library within small and large organizations
necessarily takes time and consideration, so this is not unexpected. The one
exception is, of course, Google, which has already deployed TensorFlow for a
variety of learning tasks \cite{drugs, phones, emails, deepmind, inception}. We
begin with a review of selected mentions of TensorFlow in literature. Then, we
discuss where and how TensorFlow is used in industry.

\subsection{In Literature}\label{sec:uses-lit}

The first noteworthy mention of TensorFlow is \cite{szegedy2016}, published by
Szegedy, Ioffe and Vanhoucke of the Google Brain Team in February 2016. In their
work, the authors use TensorFlow to improve on the \emph{Inception} model
\cite{inception}, which achieved best performance at the 2014 ImageNet
classification challenge. The authors report a 3.08\% top-5 error on the
ImageNet test set.

In \cite{drugs}, Ramsundar et al. discuss massively ``multitask networks for
drug discovery'' in a joint collaboration work between Stanford University and
Google, published in early 2016. In this paper, the authors employ deep neural
networks developed with TensorFlow to perform virtual screening of potential
drug candidates. This is intended to aid pharmaceutical companies and the
scientific community in finding novel medication and treatments for human
diseases.

August and Ni apply TensorFlow to create recurrent neural networks for
optimizing dynamic decoupling, a technique for suppressing errors in quantum
memory \cite{august}. With this, the authors aim to preserve the coherence of
quantum states, which is one of the primary requirements for building universal
quantum computers.

Lastly, \cite{barzdins2016} investigates the use of sequence to sequence neural
translation models for natural language processing of multilingual media
sources. For this, Barzdins et al. use TensorFlow with a sliding-window approach
to character-level English to Latvian translation of audio and video
content. The authors use this to segment TV and radio programs and cluster
individual stories.

\subsection{In Industry}\label{sec:uses-industry}

Adoption of TensorFlow in industry is currently limited only to Google, at least
to the extent that is publicly known. We have found no evidence of any other
small or large corporation stating its use of TensorFlow. As mentioned, we link
this to TensorFlow's late release. Moreover, it is obvious that many companies
would not make their machine learning methods public even if they do use
TensorFlow. For this reason, we will review uses of TensorFlow only within
Google, Inc.

Recently, Google has begun augmenting its core search service and accompanying
\emph{PageRank} algorithm\cite{pagerank} with a system called
\emph{RankBrain}\cite{rankbrain}, which makes use of TensorFlow. RankBrain uses
large-scale distributed deep neural networks for search result
ranking. According to \cite{rankbrain}, more than 15 percent of all search
queries received on www.google.com are new to Google's system. RankBrain can
suggest words or phrases with similar meaning for unknown parts of such queries.

Another area where Google applies deep learning with TensorFlow is smart email
replies \cite{emails}. Google has investigated and already deployed a feature
whereby its email service \emph{Inbox} suggests possible replies to received
email. The system uses recurrent neural networks and in particular LSTM modules
for sequence-to-sequence learning and natural language understanding. An encoder
maps a corpus of text to a ``thought vector'' while a decoder synthesizes
syntactically and semantically correct replies from it, of which a selection is
presented to the user.

In \cite{phones} it is reported how Google employs convolutional neural networks
for image recognition and automatic text translation. As a feature integrated
into its Google Translate mobile app, text in a language foreign to the user is
first recognized, then translated and finally rendered on top of the original
image. In this way, for example, street signs can be translated. \cite{phones}
notes especially the challenge of deploying such a system onto low-end phones
with slow network connections. For this, small neural networks were used and
trained to discover only the most essential information in order to optimize
available computational resources.

Lastly, we make note of the decision of Google DeepMind, an AI division within
Google, to move from Torch7 to TensorFlow \cite{deepmind}. A related source,
\cite{tpu}, states that DeepMind made use of TensorFlow for its
\emph{AlphaGo}\footnote{https://deepmind.com/alpha-go} model, alongside Google's
newly developed Tensor Processing Unit (TPU), which was built to integrate
especially well with TensorFlow. In a correspondence of the authors of this
paper with a member of the Google DeepMind team, the following four reasons were
revealed to us as to why TensorFlow is advantageous to DeepMind:

\begin{enumerate}
\item TensorFlow is included in the Google Cloud
  Platform\footnote{https://cloud.google.com/compute/}, which enables easy
  replication of DeepMind's research.
\item TensorFlow's support for TPUs.
\item TensorFlow's main interface, Python, is one of the core languages at
  Google, which implies a much greater internal tool set than for Lua.
\item The ability to run TensorFlow on many GPUs.
\end{enumerate}


\section{Conclusion}\label{sec:conclusion}

We have discussed TensorFlow, a novel open source deep learning library based on
computational graphs. Its ability to perform fast automatic gradient
computation, its inherent support for distributed computation and specialized
hardware as well as its powerful visualization tools make it a very welcome
addition to the field of machine learning. Its low-level programming interface
gives fine-grained control for neural net construction, while abstraction
libraries such as TFLearn allow for rapid prototyping with TensorFlow.  In the
context of other deep learning toolkits such as Theano or Torch, TensorFlow adds
new features and improves on others. Its performance was inferior in comparison
at first, but is improving with new releases of the library.

We note that very little investigation has been done in literature to evaluate
TensorFlow's qualities with respect to distributed execution. We esteem this one
of its principle strong points and thus encourage in-depth study by the academic
community in the future.

TensorFlow has gained great popularity and strong support in the open-source
community with many third-party contributions, making Google's move a sensible
decision already. We believe, however, that it will not only benefit its parent
company, but the greater scientific community as a whole; opening new doors to
faster, larger-scale artificial intelligence.



\appendices

\section{}\label{app:code}

\begin{lstlisting}
#!/usr/bin/env python
# -*- coding: utf-8 -*-
""" A one-hidden-layer-MLP MNIST-classifier. """

from __future__ import absolute_import
from __future__ import division
from __future__ import print_function

# Import the training data (MNIST)
from tensorflow.examples.tutorials.mnist import input_data

import tensorflow as tf

# Possibly download and extract the MNIST data set.
# Retrieve the labels as one-hot-encoded vectors.
mnist = input_data.read_data_sets("/tmp/mnist", one_hot=True)

# Create a new graph
graph = tf.Graph()

# Set our graph as the one to add nodes to
with graph.as_default():

    # Placeholder for input examples (None = variable dimension)
    examples = tf.placeholder(shape=[None, 784], dtype=tf.float32)
    # Placeholder for labels
    labels = tf.placeholder(shape=[None, 10], dtype=tf.float32)

    weights = tf.Variable(tf.truncated_normal(shape=[784, 10], stddev=0.1))
    bias = tf.Variable(tf.constant(0.1, shape=[10]))

    # Apply an affine transformation to the input features
    logits = tf.matmul(examples, weights) + bias
    estimates = tf.nn.softmax(logits)

    # Compute the cross-entropy
    cross_entropy = -tf.reduce_sum(labels * tf.log(estimates),
                                   reduction_indices=[1])
    # And finally the loss
    loss = tf.reduce_mean(cross_entropy)

    # Create a gradient-descent optimizer that minimizes the loss.
    # We choose a learning rate of 0.01
    optimizer = tf.train.GradientDescentOptimizer(0.5).minimize(loss)

    # Find the indices where the predictions were correct
    correct_predictions = tf.equal(
        tf.argmax(estimates, dimension=1),
        tf.argmax(labels, dimension=1))
    accuracy = tf.reduce_mean(tf.cast(correct_predictions, tf.float32))

with tf.Session(graph=graph) as session:
    tf.initialize_all_variables().run()
    for step in range(1001):
        example_batch, label_batch = mnist.train.next_batch(100)
        feed_dict = {examples: example_batch, labels: label_batch}
        if step % 100 == 0:
            _, loss_value, accuracy_value = session.run(
              [optimizer, loss, accuracy],
              feed_dict=feed_dict
            )
            print("Loss at time {0}: {1}".format(step, loss_value))
            print("Accuracy at time {0}: {1}".format(step, accuracy_value))
        else:
            optimizer.run(feed_dict)
\end{lstlisting}








\bibliographystyle{IEEEtran}
\bibliography{bib/IEEEabrv,bib/references}
%
%
%


\end{document}